\documentclass[letterpaper]{article} 
\usepackage[submission]{aaai25}  
\usepackage{times}  
\usepackage{helvet}  
\usepackage{courier}  
\usepackage[hyphens]{url}  
\usepackage{graphicx} 
\usepackage{natbib}  
\usepackage{caption} 
\usepackage{algorithm}
\usepackage{algorithmic}

\usepackage{soul}
\usepackage{amssymb}
\usepackage{tikz}
\usepackage[utf8]{inputenc}
\usepackage{graphicx}
\usepackage{amsmath}
\usepackage{booktabs}
\usepackage{algorithm}
\usepackage{algorithmic}
\usepackage{subcaption}
\usepackage{array}
\usepackage{textcomp}
\usepackage{verbatim}
\usepackage{xcolor}
\usepackage{enumitem}
\usepackage{arydshln}
\usepackage{fancyhdr}
\usepackage{varwidth}
\usepackage{arydshln}
\usepackage{multirow}
\usepackage{mathtools}
\usepackage{lipsum}
\usepackage{sidecap}

\usepackage{caption}
\usepackage[accsupp]{axessibility}  
\usepackage{kotex}
\usepackage{enumitem}

\definecolor{caribbeangreen}{rgb}{0.0, 0.6, 0.6}

\newcommand{\remove}[1]{}

\DeclareMathOperator*{\argmin}{arg\,min}

\usepackage{newfloat}
\usepackage{listings}
\DeclareCaptionStyle{ruled}{labelfont=normalfont,labelsep=colon,strut=off} 
\floatstyle{ruled}
\newfloat{listing}{tb}{lst}{}
\floatname{listing}{Listing}
\newcommand{\sys}{\textsf{BankTweak}}

\title{BankTweak: Adversarial Attack against Multi-Object Trackers by Manipulating Feature Banks}
\author{
    Woojin Shin\textsuperscript{\rm 1},
    Donghwa Kang\textsuperscript{\rm 2}, 
    Daejin Choi\textsuperscript{\rm 3}, 
    Brent Kang\textsuperscript{\rm 2}, 
    Jinkyu Lee\textsuperscript{\rm 4}, 
    Hyeongboo Baek\textsuperscript{\rm 1} 
}
\affiliations{
    \textsuperscript{\rm 1}University of Seoul \\
    \textsuperscript{\rm 2}Korea Advanced Institute of Science and Technology(KAIST) \\
    \textsuperscript{\rm 3}Incheon National University \\
    \textsuperscript{\rm 4}Sungkyunkwan University \\
}

\usepackage{bibentry}

\begin{document}

\maketitle

\begin{abstract}

Multi-object tracking (MOT) aims to construct moving trajectories for objects, and modern multi-object trackers mainly utilize the tracking-by-detection methodology. 
Initial approaches to MOT attacks primarily aimed to degrade the detection quality of the frames under attack, thereby reducing accuracy only in those specific frames, highlighting a lack of \textit{efficiency}. 
To improve efficiency, recent advancements manipulate object positions to cause persistent identity (ID) switches during the association phase, even after the attack ends within a few frames. 
However, these position-manipulating attacks have inherent limitations, as they can be easily counteracted by adjusting distance-related parameters in the association phase, revealing a lack of \textit{robustness}.
In this paper, we present \sys{}, a novel adversarial attack designed for MOT trackers, which features efficiency and robustness.
\sys{} focuses on the feature extractor in the association phase and reveals vulnerability in the Hungarian matching method used by feature-based MOT systems. 
Exploiting the vulnerability, \sys{} induces persistent ID switches (addressing \textit{efficiency}) even after the attack ends by strategically injecting altered features into the feature banks without modifying object positions (addressing \textit{robustness}).
To demonstrate the applicability, we apply \sys{} to three multi-object trackers (DeepSORT, StrongSORT, and MOTDT) with one-stage, two-stage, anchor-free, and transformer detectors.
Extensive experiments on the MOT17 and MOT20 datasets show that our method substantially surpasses existing attacks, exposing the vulnerability of the tracking-by-detection framework to \sys{}.

\end{abstract}

\section{Introduction}
\label{sec:intro}

Multi-object tracking (MOT) is a common perception task aimed at constructing the motion trajectories of objects across consecutive frames. 
The modern DNN-based \textit{tracking-by-detection} paradigm involves first identifying all objects of interest in each frame (\textit{detection}), then using appearance and motion cues to link these detections to existing trajectories (\textit{association}). 
For association, a CNN-based model (e.g., OSNet~\cite{ZYC19}) extracts features from each detected object, storing them in a \textit{feature bank} if matched via feature-based matching or motion-based IoU (Intersection over Union) matching. 
Initially, feature-based matching pairs objects with the highest feature similarity across frames using the feature bank, followed by motion-based IoU matching for unmatched objects.

\begin{figure}[t]
    \centering
    \includegraphics[width=1\linewidth,height=0.33\linewidth]{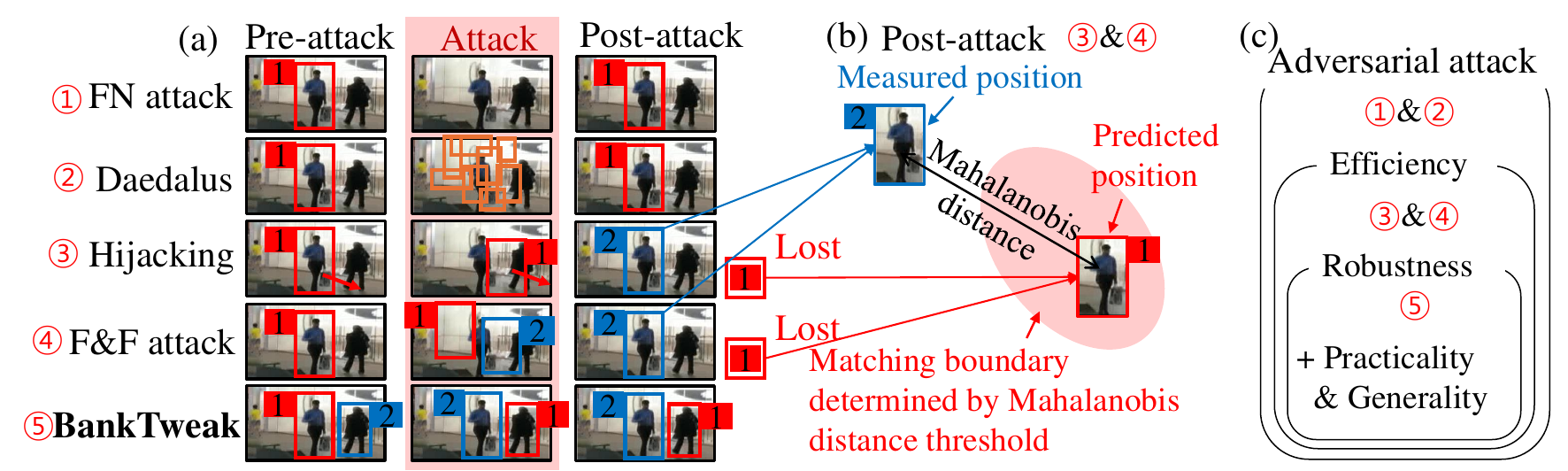}
    \caption{(a) Comparison between existing adversarial attacks \textcircled{1}--\textcircled{4} and \sys{} \textcircled{5}, (b) the principles behind ID switch induction in  Hijacking \textcircled{3} and the F\&F attack \textcircled{4}, and (c) the diagram presenting the features of adversarial attacks.}
    \label{fig:various_trackers}
    \vspace{-0.4cm}
\end{figure}

Adversarial attacks on MOT systems in the tracking-by-detection framework can be divided into two approaches: i) targeting the detection phase to degrade detection performance by generating false negatives and false alarms, and ii) disrupting the association phase, causing identity (ID) switches among detected objects by manipulating object positions.
Category i) attacks, such as the false negative (FN) attack~\cite{LSF17} (\textcircled{1} in Fig.~\ref{fig:various_trackers})
and Daedalus~\cite{wang2021daedalus} (\textcircled{2}), generate false negatives and false alarms in the attacked frames, respectively.
However, since these methods focus solely on the detection phase, they reduce accuracy only in the targeted frames and do not affect subsequent frames, resulting in a lack of \textit{efficiency}. 
As shown in Fig.~\ref{fig:various_trackers}(a), in FN attack and Daedalus scenarios, the object with ID 1 in the pre-attack frame retains ID 1 in the post-attack frame, with no drop in post-attack accuracy.

In Category ii), Hijacking~\cite{JLS20} (\textcircled{3}) manipulates the Kalman filter~\cite{KRE60} by shifting the detection box away from the correct velocity, potentially triggering ID switches. 
The F\&F attack~\cite{ZYL23} (\textcircled{4}) removes the target object and surrounds it with false alarms, causing persistent ID switches even after the attack ends.
As shown in Fig.~\ref{fig:various_trackers}(b), Category ii) attacks move the original object's position in the pre-attack frame, creating a large Mahalanobis distance~\cite{DJD00} from the original object in the post-attack frame. This distance exceeds the matching boundary, causing the system to misidentify them as different objects and assign a new ID. 
The matching boundary, based on the Mahalanobis distance, determines if objects can be candidates for matching. If within the boundary, feature-based and IoU-based matching are conducted sequentially.
Thus, Category ii) faces inherent limitations: increasing the Mahalanobis distance threshold (expanding the matching boundary) in prediction models, such as the Kalman filter, can easily counteract these attacks with minimal accuracy drop, revealing a lack of \textit{robustness}, to be demonstrated in Sec.~\ref{subsec:ablation}. 
Therefore, the limitations of Categories i) and ii) bring the necessity for an advanced category of attacks that can induce persistent ID switches (addressing \textit{efficiency}) and are position-independent (addressing \textit{robustness}).

In this paper, we propose \sys{}, a novel adversarial attack designed for MOT systems, which features efficiency and robustness. 
\sys{} focuses on the feature extractor in the association phase and performs in two steps where desired generated features are injected into the feature bank of the selected object pair without triggering an ID switch (Step 1 termed as groundwork), followed by using these preparations to induce a permanent ID switch (Step 2 called ID switch).
To be detailed in Sec.~\ref{sec:method} with Figs.~\ref{fig:combined}(b) and (c), inducing persistent ID switches in a position-independent way is highly challenging due to the inherent properties of feature banks, but \sys{} overcomes this challenge by exploiting the vulnerability of the Hungarian matching method used by feature-based MOT systems. 
Note that \sys{} enables simultaneous attacks on multiple object pairs in a single frame.

As additional key advantages, unlike the F\&F attack increasing the risk of detection by defenders due to many false alarms,  
\sys{} creates no false positives, thereby enhancing \textit{practicality}. 
Moreover, unlike the Hijacking attack requiring model-specific design like Kalman filter information, \sys{} does not rely on such specifics, ensuring greater \textit{generality}. 
Fig.~\ref{fig:various_trackers}(c) represents the characteristics of MOT attacks, including \sys{}. 
Importantly, \sys{} is not tailored to a specific tracker, making it compatible with \textit{most} DNN-based tracking-by-detection methods that use a feature bank and a two-stage association approach (i.e., feature-based and subsequent IoU-based).

To demonstrate the applicability, \sys{} is applied to three prominent multi-object trackers (DeepSORT~\cite{WBP17}, StrongSORT~\cite{DZS23}, and MOTDT~\cite{CAZ18}) with one-stage (YOLOX~\cite{GLW21}), two-stage (Faster-RCNN~\cite{RHG15}), anchor-free (FoveaBox~\cite{KSL20}) and transformer (DETR~\cite{XSL21}) detectors.
Comprehensive experiments conducted on the MOT17~\cite{MLL16} and MOT20~\cite{DRM20} datasets demonstrate that our approach significantly outperforms existing attacks, revealing the vulnerability of the tracking-by-detection framework to \sys{}.

Our contributions are as follows.

\begin{itemize}
    \item We propose a novel adversarial attack characterized by efficiency and robustness, with benefits of practicality and generality.
    \item We show the applicability of our approach by deploying it to three multi-object trackers (i.e., DeepSORT, StrongSORT, and MOTDT) with one-stage, two-stage, anchor-free, and transformer detectors.
    \item Extensive experiments on public datasets show that our method substantially surpasses existing attacks, exposing the vulnerability of the tracking-by-detection framework to \sys{}.
\end{itemize}

\section{Related work}
\label{sec:relatedwork}

Several adversarial attacks such as FGSM~\cite{goodfellow2014explaining} and PGD~\cite{madry2017towards} were proposed to efficiently address perturbations. 
Recently, extensive research has focused on adversarial attacks for various visual recognition tasks, including detection~\cite{wang2021daedalus, xie2017adversarial}, tracking~\cite{yan2020cooling, JLS20}, and semantic segmentation~\cite{xie2017adversarial}. Some studies~\cite{xu2020adversarial, chen2019shapeshifter, zhong2022shadows, duan2021adversarial} have even brought adversarial attacks to the real world.
In the realm of single object tracking, which is closely related to MOT, various attacks have been proposed based on different intuitions~\cite{xu2020adversarial, chen2020one}. Additionally, detection attacks have induced various malfunctions, such as missed detections~\cite{LSF17} or false alarms~\cite{wang2021daedalus}. However, these methods have shown limited effectiveness in attacking multi-object trackers due to mission gaps.
Recently, Jia et al.~\cite{JLS20} introduced MOT attacks focusing on deceiving the Kalman filter~\cite{KRE60} inside multi-object trackers. To enhance effectiveness, they adopted an iterative method for spreading perturbations to tackle disturbances. Concurrently, the F\&F attack~\cite{ZYL23} achieves ID switches in tracked objects by simultaneously triggering missed detections and false alarms.

%
%

\section{Method}
\label{sec:method}


\subsection{Attack formulation}
\label{subsec:att_formulation}
 



We consider a tracking-by-detection MOT system, consisting of detection and association phases, discussed in Sec.~\ref{sec:intro}. 
\sys{} operates under the assumption of a \textit{white box} attack, with the attacker knowing the detector and feature extractor models for the iterative execution of detection and feature extraction across attack frames.
\sys{} only needs five frames for an attack without any false alarms (practicality), 
and does not necessitate motion prediction of objects (generality).

Consider an input video comprised of $N$ sequential RGB frames $I \in \mathbb{R}^{W \times H \times 3}$, represented as $\mathbb{V} = \{I_{1}, I_{2}, \cdots, I_{N}\}$.  
We target a sequence of five consecutive frames starting from the $(t+1)$-th frame (for $1 \le t$) for our attack, denoted by $\mathbb{S} = \{I_{t+1}, I_{t+2},$ $I_{t+3}, I_{t+4}, I_{t+5}\}$. 
Let $\tilde{I}_{t}$ be a frame created by adding a perturbation $\delta$ to $I_{t}$. 
By incorporating perturbations into every frame in $\mathbb{S}$, we generate $\tilde{\mathbb{S}} = \{\tilde{I}_{t+1}, \tilde{I}_{t+2},  \tilde{I}_{t+3}, \tilde{I}_{t+4}, \tilde{I}_{t+5}\}$ (for $t+5 < N$). 
Substituting $\mathbb{S}$ in $\mathbb{V}$ with $\tilde{\mathbb{S}}$ yields $\tilde{\mathbb{V}} = \{I_{1}, I_{2}, \cdots, I_{t}, \tilde{I}_{t+1}, \tilde{I}_{t+2}, 
\cdots, 
\tilde{I}_{t+5}, I_{t+6}$, $I_{t+7}$ $ \cdots, I_{N}\}$.

\begin{algorithm}[t!]
\caption{\sys{} attack}
\textbf{Input:} target frame sequence $\mathbb{S}$, object detector $D(\cdot)$, feature extractor $E(\cdot, \cdot)$
\raggedright
\textbf{Output:} perturbed frame sequence $\tilde{\mathbb{S}}$\\
\begin{algorithmic}[1]
    \STATE $\tilde{\mathbb{S}} =$ [ ]
    \FOR{$I$ from $I_{t+1}$ to $I_{t+5}$ in $\mathbb{S}$}              
        \STATE $\mathbb{F}^* = E(D(I), I)$ 
        ~~~~~~~~/* get targeted feature set and loss function (Sec.~\ref{subsec:mechanism}) */
        \STATE $\mathbb{F}, \mathcal{L} \xleftarrow{}$ get\_targeted\_features($\mathbb{F}^*$) 
        ~~/* solve perturbation with Eq.~\eqref{eq:pgd}--~\eqref{eq:iter_delta} */
        \STATE $\tilde{I} \xleftarrow{}$ solve\_perturbation($\mathbb{F}, \mathcal{L}, I, D(\cdot), E(\cdot, \cdot)$)
        \STATE $\tilde{\mathbb{S}}$.append($\tilde{I}$)
    \ENDFOR
    \RETURN $\tilde{\mathbb{S}}$
\end{algorithmic}
\label{alg:sys_alg}
\end{algorithm}

For the given target input frame $I$, the detector $D(\cdot|\theta_{D})$ parameterized by $\theta_{D}$, feature extractor $E(\cdot, \cdot|\theta_{E})$ parameterized by $\theta_{E}$, and target features $\mathbb{F}$, \sys{} finds perturbation $\delta$ formulated by

\begin{equation}
    \delta = \argmin_{\delta, ||\delta||_{\infty} < \epsilon} \mathcal{L}(E(D(I+\delta|\theta_{D}), I+\delta|\theta_{E}), \mathbb{F}), 
    \label{eq:pgd}
\end{equation}


\noindent
where $D(\cdot|\theta_{D})$ processes an input frame $I$ to identify the detected object set $\mathbb{O}$ and $E(\cdot, \cdot|\theta_{E})$ extracts feature set $\mathbb{F}^*$ from $\mathbb{O}$. 
\sys{} uses PGD to iteratively derive a perturbation $\delta$ that minimizes $\mathcal{L}$ under an $\ell_{\infty}$-norm constrain ($\mathcal{L}$ will be 
further detailed in Sec.~\ref{subsec:perturbation}), which is 

\begin{align}
    \delta^{r+1} &= \text{clip}_{[-\epsilon, \epsilon] \cap [-I, 1-I]} \nonumber \\
    &(\delta^{r} + \alpha \text{sgn}(\nabla_{\delta}\mathcal{L}(E(D(I+\delta|\theta_{D}), I+\delta|\theta_{E}), \mathbb{F}))), 
    \label{eq:iter_delta} 
\end{align}


\noindent
where $\alpha$ and $\epsilon$ represent the amount of change per pixel and the maximum change allowed, respectively, while $\nabla$ and sgn($\cdot$) are functions for performing the gradient operation and extracting the sign of the gradient, respectively. 
\sys{} initializes the first perturbation $\delta$ to zero and iterates $R$ times to compute the final $\delta$. During these iterations, $\delta$ must adhere to an $\ell_{\infty}$-norm constraint, ensuring the perturbed frame $\tilde{I}$ remains within the [0, 1] range.



Alg.~\ref{alg:sys_alg} outlines the attack process of \sys{}. 
It takes as input the target frame sequence $\mathbb{S}$, object detector $D(\cdot)$, and feature extractor $E(\cdot, \cdot)$, producing the perturbed frame sequence $\tilde{\mathbb{S}}$ as output. 
For each input frame $I$, \sys{} performs the detection to obtain the object set $\mathbb{O}$ and then conducts feature extraction based on $\mathbb{O}$ to extract the feature set $\mathbb{F}^*$ (Line 3). 
Subsequently, it determines the designated $\mathbb{F}$ and $\mathcal{L}$ for each attack frame based on $\mathbb{F}^*$ (Line 4). 
Utilizing the derived $\mathbb{F}$, $\mathcal{L}$, and the models $D(\cdot)$ and $E(\cdot, \cdot)$, it computes the perturbed frame $\tilde{I}$ using Eqs.~\eqref{eq:pgd} and~\eqref{eq:iter_delta} (Line 5), which is then added to $\tilde{\mathbb{S}}$. 
The detector $D(\cdot)$ is used for cropping the detected object from the input image after performing detection, and the perturbation is determined through the model $E(\cdot, \cdot)$ (Line 5). 
This procedure is repeated for the length of the input frame sequence $\mathbb{S}$, which is five (i.e., from $I_{t+1}$ to $I_{t+5}$), and ultimately returns the perturbed frame sequence $\tilde{\mathbb{S}}$ (Line 8). 

\begin{figure}[t!]
    \centering
    \includegraphics[width=1\linewidth,height=0.33\linewidth]{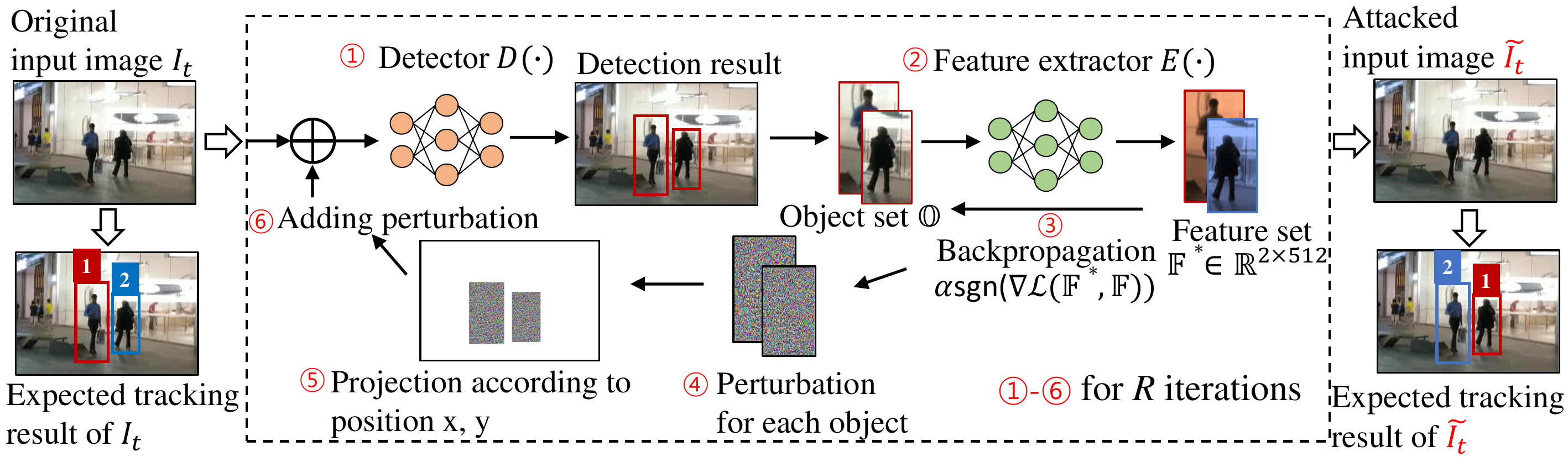}
    \caption{Generating perturbed features by iterating \textcircled{1}--\textcircled{6} to induce ID switch between two objects in \sys{}: a focus on the feature extractor during the association phase.}
    \label{fig:iter_attack}
    \vspace{-0.3cm}
\end{figure}

\begin{figure*}[t]
    \centering
    \includegraphics[width=0.92\linewidth]{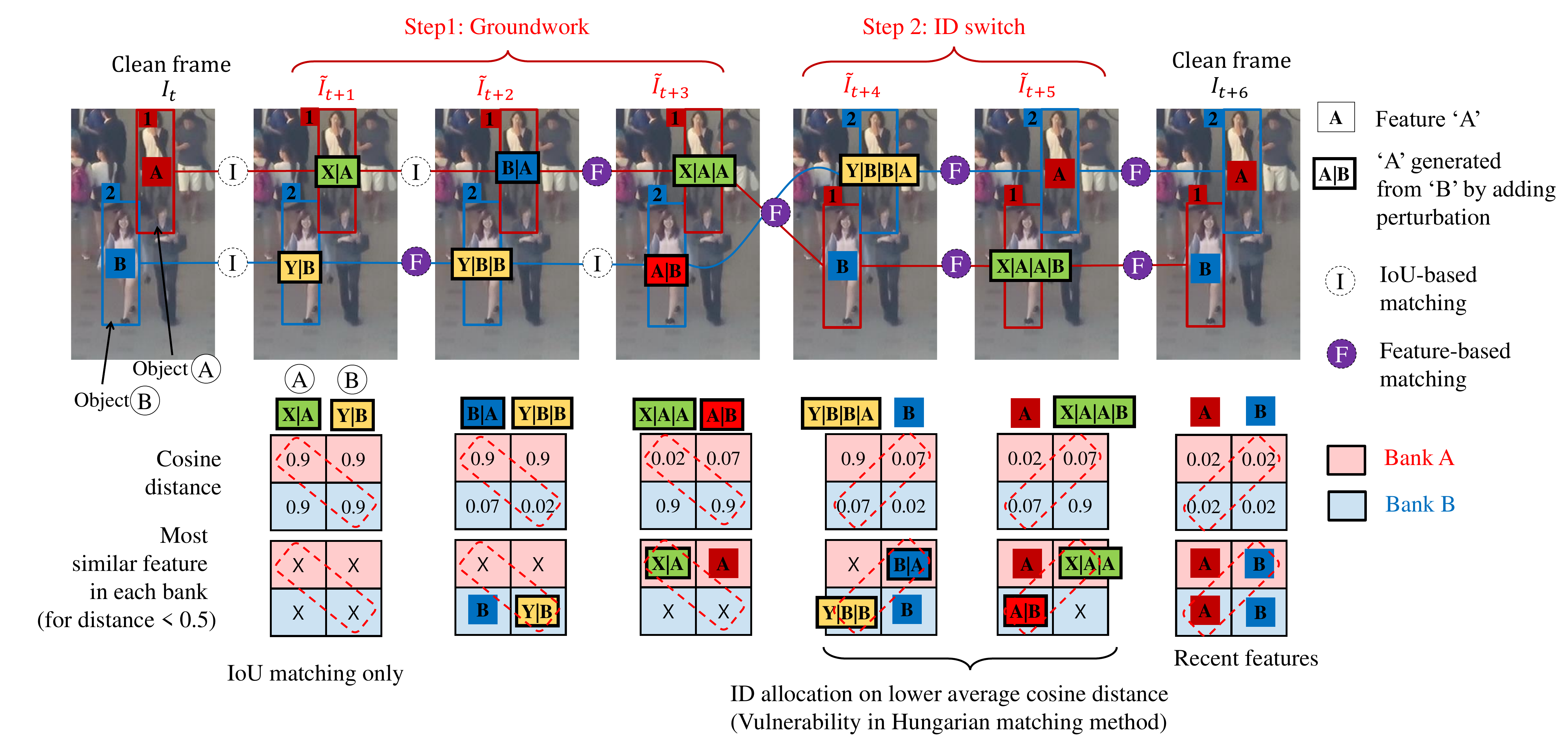}
    \caption{Overall process by which \sys{} induces an ID switch between a pair of objects {\textcircled{A}} and {\textcircled{B}}, across five frames from $\tilde{I}_{t+1}$ to $\tilde{I}_{t+5}$; desired generated features are injected into the feature bank in $\tilde{I}_{t+1}$--$\tilde{I}_{t+3}$ (Step 1), and a permanent ID switch incurs by exploiting vulnerability in Hungarian matching method in $\tilde{I}_{t+4}$--$\tilde{I}_{t+5}$ (Step 2).}
    \label{fig:main_method}
    \vspace{-0.2cm}
\end{figure*}

\subsection{\sys{} mechanism}
\label{subsec:mechanism}


Fig.~\ref{fig:iter_attack} presents the overall process to generate perturbations in \sys{}.
To create perturbed features, \sys{} begins with detecting objects in the input image $I_t$ to identify the object set $\mathbb{O}$ (\textcircled{1}). 
These objects are cropped and resized from the input image, aggregated into a batch, and then processed through the feature extractor to extract the feature set $\mathbb{F}^*$ (\textcircled{2}). 
A comparison of the similarity between $\mathbb{F}^*$ and the intended target feature set $\mathbb{F}$ is made, with the loss calculated using the predefined loss function $\mathcal{L}$ (\textcircled{3}). 
The loss determined for each object is used to identify specific perturbations (\textcircled{4}), 
which take into account the objects' coordinates and are then projected onto the input image (\textcircled{5}), effectively integrating these perturbations into the original image (\textcircled{6}). 
This procedure is iterated $R$ times.

The primary objective of \sys{} is to switch the IDs of target objects \textcircled{A} and \textcircled{B}, ensuring these changes remain constant, even after the completion of the attack. This process unfolds in two steps.
(i) Initially, \sys{} systematically injects perturbed features into the feature banks of \textcircled{A} and \textcircled{B}, without initiating an ID switch. 
This preparatory action lays the groundwork for the next step.
(ii) Subsequently, leveraging the altered feature banks established in (i), \sys{} executes the ID switch for \textcircled{A} and \textcircled{B}, effectively achieving the intended consistent ID switch even after the attack.
Fig.~\ref{fig:main_method} presents the overall process by which \sys{} induces an ID switch between a pair of objects 
{\textcircled{A}} and {\textcircled{B}.
For a clean frame $I_t$, features $A$ and $B$ are extracted from objects {\textcircled{A}} and {\textcircled{B}, respectively, and are assigned ID 1 and ID 2. 
These features are then stored in the feature banks of their corresponding objects.  
Consider $A|B$ as feature $A$ generated from $B$ through an adversarial example mechanism (e.g., PGD~\cite{madry2017towards}), which appears as $B$ to humans but is identified as $A$ by the deployed model. 
\sys{} selects the object pair {\textcircled{A}} and {\textcircled{B} for an ID switch in each frame $I_t$, fundamentally choosing {\textcircled{A}} and {\textcircled{B} randomly without awareness of the Mahalanobis distance threshold, thereby satisfying generality. 

\textbf{Step 1: Groundwork.} It performs the following for the first three attacked frames:
\begin{itemize}
    \item [] $\tilde{I}_{t+1}$: Define $X$ and $Y$ as the dummy features that exhibit a significantly large cosine distance from $A$ and $B$, ensuring they are distinctly different. By definition, $X|A$ and $Y|B$ have a high cosine distance (e.g., 0.9) from $A$ and $B$, respectively, and are injected into the feature banks of \textcircled{A} and \textcircled{B} through IoU matching; it is assumed that features are only considered for feature-based matching when they have a cosine distance of 0.2 or less (called cosine distance threshold).
    \item [] $\tilde{I}_{t+2}$: It places $B|A$ into \textcircled{A}'s feature bank and $Y|B|B$ into \textcircled{B}'s feature bank, leveraging the very low cosine distance between $Y|B|B$ and $Y|B$ (e.g., 0.02 in Fig.~\ref{fig:main_method}) once it is inserted into \textcircled{B}'s bank in $\tilde{I}_{t+1}$. Being generated from $A$, $B|A$ exhibits a relatively low cosine distance (e.g., 0.07 in Fig.~\ref{fig:main_method}).
    \item [] $\tilde{I}_{t+3}$: It places $A|B$ into \textcircled{B}'s feature bank and $X|A|A$ into \textcircled{A}'s feature bank using the similar property in $\tilde{I}_{t+2}$.
\end{itemize}

\noindent
Fig.~\ref{fig:combined}(a) presents our experimental result for two distinct scenarios, each featuring varying converging cosine distances when a source feature is subjected to up to 150 perturbations to derive a specific target feature, as outlined in Eq.~\eqref{eq:make_sim}. 
For instance, when producing $Y|B|B|A$ (in $\tilde{I}_{t+4}$), the source feature is $A$, targeting the feature $Y|B|B$ (in $\tilde{I}_{t+2}$). Conversely, $Y|B|B$ is derived from $B$, resulting in a cosine distance of 0.07 between $Y|B|B|A$ and $Y|B|B$ due to the disparity in their source features.
On the other hand, for the production of $Y|B|B$ (in $\tilde{I}_{t+2}$), $B$ acts as the source feature with $Y|B$ (in $\tilde{I}_{t+1}$) as the target. 
Here, $Y|B$, created from the same source $B$, leads to a minimal cosine distance of 0.02 between $Y|B|B$ and $Y|B$.
It might seem straightforward to induce an ID switch by injecting $B|A$ and $A|B$ into the feature banks of {\textcircled{A}} and {\textcircled{B} in Step 1. 
However, Fig.~\ref{fig:combined}(b) illustrates that because the feature banks of {\textcircled{A}} and {\textcircled{B} already include $A$ and $B$, any ID switch in $\tilde{I}_{t+1}$ reverts to the original IDs by $\tilde{I}_{t+2}$ post-attack.

\textbf{Step 2: ID switch.} This step involves the following for the next two frames:
\begin{itemize}
\item [] $\tilde{I}_{t+4}$: It places $Y|B|B|A$ into \textcircled{A}'s feature bank without creating any perturbation for \textcircled{B}. 
 $B$ in $\tilde{I}_{t+4}$ demonstrates a significantly low cosine distance (i.e., 0.02) to $B$ in $I_t$, whereas $Y|B|B|A$ exhibits a high cosine distance (i.e., 0.9) with the features in \textcircled{A}'s bank. 
 Given the cosine distances of 0.07 between $Y|B|B|A$ and $Y|B|B$, and 0.07 between $B$ and $B|A$, the Hungarian algorithm~\cite{KHW55} allocates IDs based on the lower average cosine distance, prompting an ID switch.
\item [] $\tilde{I}_{t+5}$: It places $X|A|A|B$ into \textcircled{B}'s feature bank, and similarly, no perturbation is created for \textcircled{A}. Similar reasoning to $\tilde{I}_{t+4}$ causes an ID switch for $\tilde{I}_{t+5}$.
\end{itemize}

\noindent
As a result, from $I_{t+6}$ onward, $A$ in \textcircled{A} matches with $A$ having ID 2 in $\tilde{I}_{t+5}$ (rather than $A$ with ID 1 in $I_t$), and $B$ in \textcircled{B} matches with $B$ having ID 1 in $\tilde{I}_{t+4}$ (rather than $B$ with ID 2 in $I_t$), thus continuous ID switches occur without further attacks. 
One might assume that directly injecting $A$ and $B$ into the feature banks of {\textcircled{A}} and {\textcircled{B} for $\tilde{I}_{t+4}$ in Step 2 is a straightforward approach to trigger an ID switch. 
However, as shown in Fig.~\ref{fig:combined}(c), such an action does not lead to an ID switch due to the pre-existing $A$ and $B$ in the respective feature banks of {\textcircled{A}} and {\textcircled{B}. 
To this end, \sys{} employs a meticulous strategy that exploits the vulnerability (i.e., ID allocation on lower average cosine distance) of the Hungarian algorithm in Step 2, based on the groundwork conducted in Step 1.

\subsection{Solving perturbations}
\label{subsec:perturbation}

\sys{} employs cosine distance to evaluate the similarity between the features of two objects, facilitating the creation of the target feature set $\mathbb{F}$ from an initial feature set $\mathbb{F}^*$, which is derived by

\begin{equation}
    \mathcal{C}(A, B) = 1 - \frac{A \cdot B}{|A||B|},
    \label{eq:cosine_dist}
\end{equation}

\noindent
where $A$ and $B$ are feature vectors of two distinct objects, each in $\mathbb{R}^{1 \times 512}$.
Eq.~\eqref{eq:cosine_dist} produces values within the [0, 2] range, with lower values denoting higher similarity and higher values indicating greater dissimilarity between the features of two objects.

For a frame $I$, which includes multiple objects, we define the extracted feature set as $\mathbb{F}^*$ and its target feature set as $\mathbb{F}$. 
For each feature $F_i^* \in \mathbb{F}^*$, $F_i \in \mathbb{F}$ represents its corresponding target feature. 
\sys{} computes the loss for each feature set $\mathbb{F}^*$, aggregates these losses, and applies the perturbation collectively. 
This process employs a specific loss function formulated as

\begin{equation} 
    \mathcal{L}^{s}(\mathbb{F}^*, \mathbb{F}) = \sum_{F_i^* \in \mathbb{F}^*, F_i \in \mathbb{F}} \mathcal{C}(F_i^*, F_i), 
    \label{eq:make_sim}
\end{equation}

\begin{equation} 
    \mathcal{L}^{d}(\mathbb{F}^*, \mathbb{F}) = - \sum_{F_i^* \in \mathbb{F}^*, F_i \in \mathbb{F}}  \mathcal{C}(F_i^*, F_i). 
    \label{eq:make_diff}
\end{equation}

For each feature $F_i^* \in \mathbb{F}^*$ and its target feature $F_i \in \mathbb{F}$, the loss $\mathcal{L}^{s}$ signifies that a lower value increases the similarity between $F_i^*$ and $F_i$. 
Conversely, a higher value of $\mathcal{L}^{d}$ decreases the similarity between $F_i^*$ and $F_i$. For instance, the goal for $\tilde{I}_{t+1}$ is to generate $X|A$ and $Y|B$, thus the target feature set $\mathbb{F} = \{F_1=A, F_2=B\}$, and $X|A$ and $Y|B$ are created to have the maximum possible cosine distance from $A$ and $B$, respectively, using Eq.~\eqref{eq:make_diff}. On the other hand, for $\tilde{I}_{t+2}$, aiming to generate $B|A$ and $Y|B|B$, the target feature set $\mathbb{F} = \{F_1=B, F_2=Y|B\}$, and $B|A$ and $Y|B|B$ are produced to be as close as possible to $A$ and $B$, respectively, using Eq.~\eqref{eq:make_sim}. 
The feature sets $\mathbb{F}^*$, target feature set $\mathbb{F}$, and the loss function for each attack frame are determined as follows.

\remove{
\small
\begin{enumerate}
    \item []$\tilde{I}_{t+1}$: $\mathbb{F}^* = \{F_1^*=A, F_2^*=B\}$ and $\mathbb{F} = \{F_1=A, F_2=B\}$ with $\mathcal{L}^{d}(\mathbb{F}^*, \mathbb{F})$,
    \item []$\tilde{I}_{t+2}$: $\mathbb{F}^* = \{F_1^*=A, F_2^*=B\}$ and $\mathbb{F} = \{F_1=B, F_2=Y|B\}$ with $\mathcal{L}^{s}(\mathbb{F}^*, \mathbb{F})$,
    \item []$\tilde{I}_{t+3}$: $\mathbb{F}^* = \{F_1^*=A, F_2^*=B\}$ and $\mathbb{F} = \{F_1=X|A, F_2=A\}$ with $\mathcal{L}^{s}(\mathbb{F}^*, \mathbb{F})$,
    \item []$\tilde{I}_{t+4}$: $\mathbb{F}^* = \{F_1^*=A\}$ and $\mathbb{F} = \{F_1=Y|B|B\}$ with $\mathcal{L}^{s}(\mathbb{F}^*, \mathbb{F})$, and
    \item []$\tilde{I}_{t+5}$: $\mathbb{F}^* = \{F_1^*=B\}$ and $\mathbb{F} = \{F_1=X|A|A\}$ with $\mathcal{L}^{s}(\mathbb{F}^*, \mathbb{F})$.
\end{enumerate}
\normalsize
}

\begin{figure}[t]
\centering
    \includegraphics[width=1\linewidth]{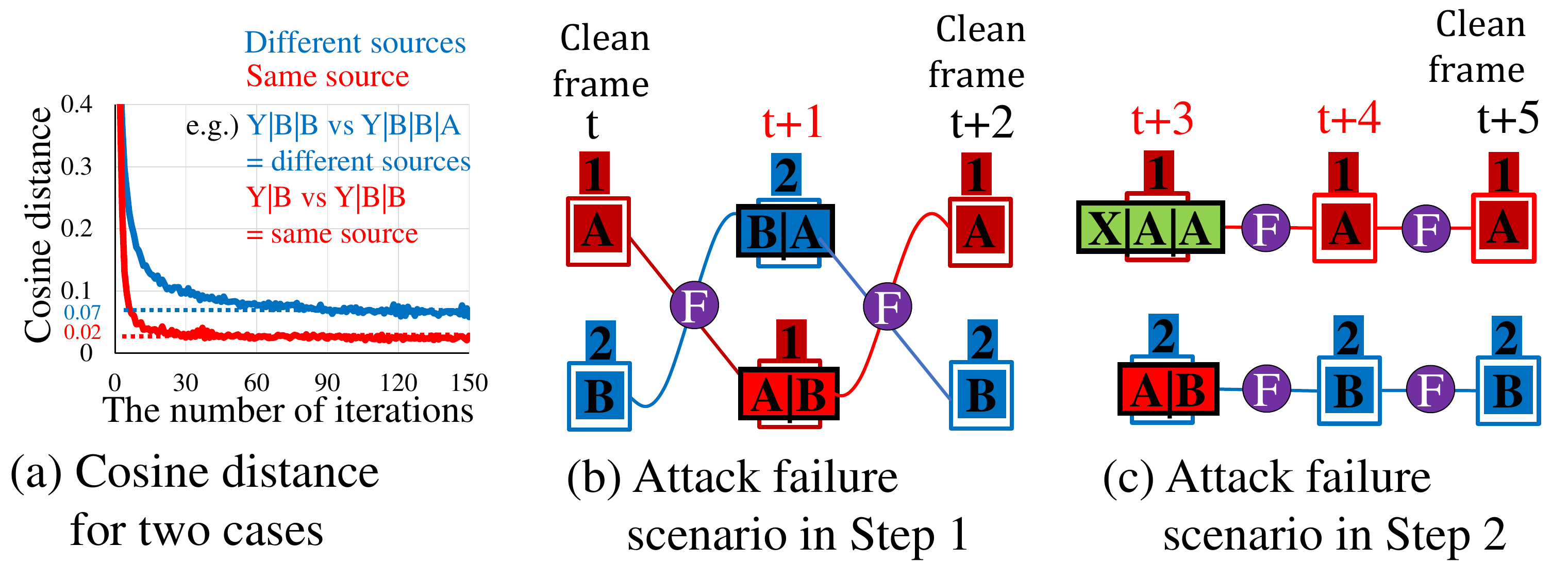}
    \caption{ (a) Comparison of cosine distance for two cases involving the same and different object sources, and (b)--(c) two attack failure scenarios of \sys{}.}
    \label{fig:combined}
    \vspace{-0.3cm}
\end{figure}

\small
\begin{tabbing}
    $\tilde{I}_{t+1}$: \= $\mathbb{F}^* = \{F_1^*=A, F_2^*=B\}$ and $\mathbb{F} = \{F_1=A, F_2=B\}$ \\
    \hspace{5em} \> with $\mathcal{L}^{d}(\mathbb{F}^*, \mathbb{F})$, \\
    $\tilde{I}_{t+2}$: \> $\mathbb{F}^* = \{F_1^*=A, F_2^*=B\}$ and $\mathbb{F} = \{F_1=B, F_2=Y|B\}$ \\
    \hspace{5em} \> with $\mathcal{L}^{s}(\mathbb{F}^*, \mathbb{F})$, \\
    $\tilde{I}_{t+3}$: \> $\mathbb{F}^* = \{F_1^*=A, F_2^*=B\}$ and $\mathbb{F} = \{F_1=X|A, F_2=A\}$ \\
    \hspace{5em} \> with $\mathcal{L}^{s}(\mathbb{F}^*, \mathbb{F})$, \\
    $\tilde{I}_{t+4}$: \> $\mathbb{F}^* = \{F_1^*=A\}$ and $\mathbb{F} = \{F_1=Y|B|B\}$ \\
    \hspace{5em} \> with $\mathcal{L}^{s}(\mathbb{F}^*, \mathbb{F})$, and \\
    $\tilde{I}_{t+5}$: \> $\mathbb{F}^* = \{F_1^*=B\}$ and $\mathbb{F} = \{F_1=X|A|A\}$ \\
    \hspace{5em} \> with $\mathcal{L}^{s}(\mathbb{F}^*, \mathbb{F})$. \\
\end{tabbing}
\normalsize

\section{Evaluation}
\label{sec:eval}

\begin{table}[t]
\caption{Attack performance comparison on YOLOX.}
\label{tab:main_exp}
\vspace{-0.3cm}
\centering
{\normalsize 
\resizebox{0.48\textwidth}{!}{%
\begin{tabular}{c|c|c|c|cc|cc}
\hline
\multicolumn{1}{c|}{Tracker} & Attacker         & IDF1 ↓         & IDsw ↑       & DetA ↓     & AssA ↓    & $\rho^{Det}$& $\rho^{ID}$ \\ \hline
\multirow{5}{*} \textbf{\underline{DS}}    & Clean            & 79.78          & 173          & 58.35      & 77.57     & 0.93x       & 0.82x          \\  
                             & FN attack        & 78.90          & 193          & 57.95      & 76.70     & 0.00x       & 0.00x          \\  
                             & Daedalus         & 74.48          & 382          & 57.08      & 71.23     & 7.18x       & 2.01x          \\  
                             & F\&F attack      & 72.03          & 493          & 56.57      & 68.11     & 3.13x       & 2.59x          \\  
                             & \textbf{\sys{}}  & \textbf{58.01} & \textbf{877} & \textbf{47.52} & \textbf{48.59} & \textbf{0.92x} & \textbf{0.82x} \\ \cline{1-8} 
\multirow{5}{*} \textbf{\underline{SS}}  & Clean            & 75.41          & 111          & 54.28      & 77.82     & 0.93x       & 0.83x         \\  
                             & FN attack        & 75.43          & 96           & 54.21      & 78.02     & 0.00x       & 0.00x          \\  
                             & Daedalus         & 74.68          & 136          & 54.10      & 76.77     & 7.18x       & 2.00x         \\  
                             & F\&F attack      & 74.26          & 158          & 53.84      & 76.02     & 3.14x       & 2.60x          \\  
                             & \textbf{\sys{}}  & \textbf{61.84} & \textbf{712} & \textbf{52.94} & \textbf{56.77} & \textbf{0.92x} & \textbf{0.84x} \\ \cline{1-8} 
\multirow{5}{*} \textbf{\underline{MD}}       & Clean            & 71.77          & 326          & 51.27      & 78.39     & 0.93x       & 0.83x         \\  
                             & FN attack        & 70.87          & 343          & 50.32      & 78.27     & 0.00x       & 0.00x          \\  
                             & Daedalus         & 69.76          & 293          & 49.19      & 76.94     & 7.18x       & 4.20x          \\  
                             & F\&F attack      & 69.68          & 271          & 48.37      & 78.18     & 3.14x       & 3.03x          \\  
                             & \textbf{\sys{}}  & \textbf{57.75} & \textbf{937} & \textbf{45.80} & \textbf{56.47}  & \textbf{0.92x} & \textbf{0.92x} \\ \hline
\end{tabular}%
}
}
\normalsize
\end{table}

\begin{table}[t!]
\caption{IDF-1 on various combinations of detectors and trackers.}
\label{tab:various_comb}
\vspace{-0.3cm}
\centering
{\normalsize 
\resizebox{0.48\textwidth}{!}{%
\begin{tabular}{c|ccc|ccc|ccc}
\hline
Detector  & \multicolumn{3}{c|}{Faster-RCNN} & \multicolumn{3}{c|}{FoveaBox} & \multicolumn{3}{c}{DETR} \\ \hline
Tracker   & \underline{DS} & \underline{SS} & \underline{MD} & \underline{DS} & \underline{SS} & \underline{MD} & \underline{DS} & \underline{SS} & \underline{MD} \\ \hline
Clean     & 66.8      & 68.8      & 67.0     & 60.6    & 63.0    & 64.0   & 52.4   & 53.3   & 54.6   \\
\textbf{BankTweak} & \textbf{56.1} & \textbf{59.9} & \textbf{53.6} & \textbf{50.4} & \textbf{55.5} & \textbf{54.0} & \textbf{48.6} & \textbf{52.5} & \textbf{48.6} \\ \hline
\end{tabular}%
}
}
\end{table}

\subsection{Experiment setting}

\textbf{Metrics.}
We compare the performance of the considered approaches regarding efficiency, robustness, practicality, and generality.
For efficiency, we utilize standard MOT accuracy metrics such as \text{IDF1}~\cite{RSZ16} and \text{HOTA}~\cite{LOD21}.
HOTA is the positive square root of the product of DetA and AssA, related to detection and association, respectively. 
DetA is the proportion of accurately detected objects, and AssA is the proportion of correctly tracked objects.
We also evaluate the number of ID Switches denoted by \text{IDsw}. 
These metrics exclude attack frames for accuracy.
For robustness, we measure accuracy by varying the Mahalanobis distance threshold.
For practicality, we measure $\rho^{\text{Det}}$ and $\rho^{\text{ID}}$ to assess the system's effectiveness in reducing new objects (mainly false alarms) during attack frames.
$\rho^{\text{Det}}$ is the ratio of the average increase in detections per attack frame to the ground truth $GT_t$, and $\rho^{\text{ID}}$ measures the increase in ID counts.
No metric is used for generality, as \sys{} is designed to satisfy generality without requiring prediction model information.

\textbf{Dataset.}
Experiments are conducted using the MOT17 and MOT20 pedestrian tracking datasets, with the experimental results for MOT20 provided in the supplementary material due to page limitations.
Each dataset is split into two halves: one for training the considered detection model and the other for evaluation.
The MOT17 and MOT20 datasets are further divided into 30-frame segments, yielding 83 and 148 segments, respectively. 
Experiments target each segment's (15–19)-th frames for attacks to accumulate features in the objects' feature banks over five frames, ensuring accurate evaluation of \sys{}'s potential effects in practical tracking applications.

\textbf{Implementation details.}
To demonstrate the applicability, \sys{} is applied to three prominent multi-object trackers (DeepSORT, StrongSORT, and MOTDT denoted by \underline{DS}, \underline{SS}, and \underline{MD}, respectively) with one-stage(YOLOX), two-stage(Faster-RCNN), anchor-free (FoveaBox) and transformer (DETR) detectors.
OSNet is considered as the feature extractor.
The feature-based matching threshold $\lambda_{app} = 0.2$ and IoU-based matching threshold $\lambda_{IoU} = 0.7$. 
Attack parameters are $\epsilon = 4/255$ and $\alpha = 1/255$.
In \sys{}, dissimilarity loss $\mathcal{L}^{d}$ succeeds when feature similarity exceeds $\lambda_{app} = 0.2$, and similarity loss $\mathcal{L}^{s}$ requires cosine distance to be less than $\lambda_{app} = 0.2$. 
Empirically, iterations for $\mathcal{L}^{d}$ are set to $R^{d} = 10$, and for $\mathcal{L}^{s}$, $R^{s} = 150$.

\begin{figure}[t!]
    \centering
    \includegraphics[width=1\linewidth]{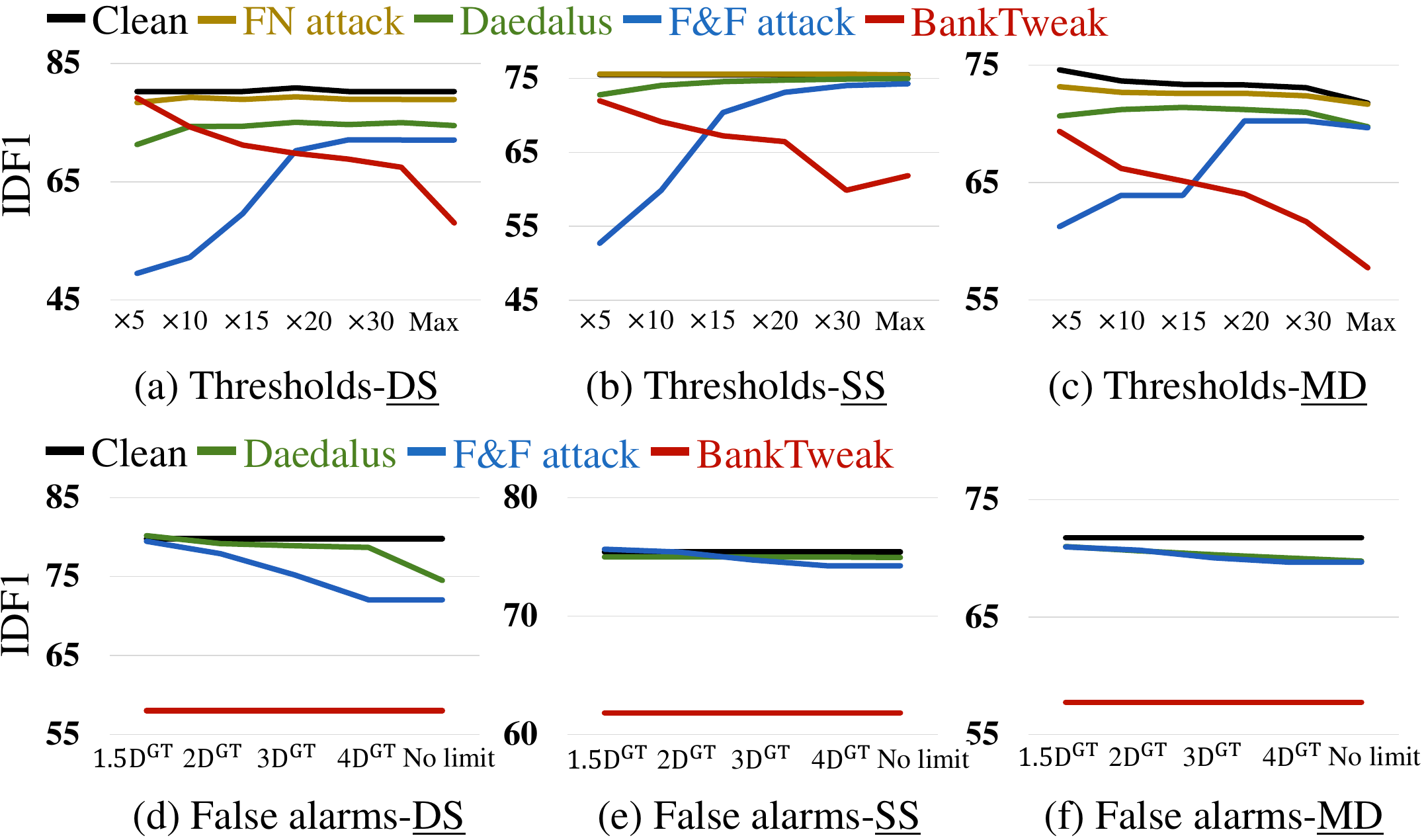}
    \caption{(a)--(c) IDF1 scores across varying Mahalanobis distance thresholds and (d)--(f) false alarms for three distinct trackers.}
    \label{fig:Mahalanobis_false_alarms}
    \vspace{-0.3cm}
\end{figure}

\subsection{Comparison to existing methods}

We evaluate our approach against three principal baselines: (i) the FN attack~\cite{LSF17}, which reduces accuracy by removing objects' bounding boxes during attack frames, causing false negatives, (ii) Daedalus~\cite{wang2021daedalus}, which induces false positives by detecting multiple objects within attack frames, and (iii) the F\&F attack~\cite{ZYL23}, which increases ID switches post-attack by creating false positives at detected object locations during attack frames. Hijacking~\cite{JLS20} is excluded as it does not meet generality. 

\textbf{Efficiency and robustness.}
Table~\ref{tab:main_exp} shows attack performance under the maximum Mahalanobis distance threshold, counteracting Hijacking and F\&F attack with minimal accuracy drop (see Figs.~\ref{fig:Mahalanobis_false_alarms}(a)--(c)) on the MOT17 dataset.
Table~\ref{tab:main_exp} (for YOLOX) demonstrates that \sys{} significantly decreases the IDF1 score across all trackers. 
For example, \sys{} lowers the IDF1 by 21.77 (i.e., 79.78 vs 58.01) for \underline{DS}. 
It also increases the association metric IDsw while decreasing AssA, alongside a reduction in the detection metric DetA. These findings verify that \sys{} is capable of efficiently causing ID switches by creating perturbations via the feature extractor and interfering with the matching procedure.
Unlike FN attack and Daedalus, \sys{} significantly affects performance by altering object IDs during attacks, with these changes persisting post-attack and resulting in a noticeable decline in IDF1 scores. The F\&F attack shifts an object's position before the attack, creating a significant Mahalanobis distance from its original location after the attack, but increasing the Mahalanobis distance threshold (expanding the matching boundary) can neutralize the attack with minimal impact on accuracy. \sys{} thrives when the matching boundary is expanded, ensuring the targeted object pair falls within this expanded boundary, enhancing performance.

\textbf{Practicality.}
\sys{} induces ID switches by altering the feature banks of individual objects, effectively reducing accuracy without generating false alarms. Table~\ref{tab:main_exp} illustrates that \sys{} keeps the object count stable during an attack, as evidenced by the $\rho^{Det}$ and $\rho^{ID}$ metrics, closely aligning with the Clean scenario. 
For instance, detected object values on \underline{DS} for Clean and \sys{} closely match ($\rho^{Det}$ at 0.93 vs 0.92), similar to the tracked object values ($\rho^{ID}$ remains at 0.82 for both). 
In contrast, Daedalus shows a marked effect on these metrics, with $\rho^{Det}$ jumping from 0.93 to 7.18 and $\rho^{ID}$ for tracked objects rising from 0.82 to 2.01, demonstrating a significant difference.

\textbf{Applicability.}
In Table~\ref{tab:main_exp}, against \underline{DS}, Daedalus and F\&F attacks reduce the IDF1 score from 79.78 to 74.48 and 72.03, respectively. 
In comparison, the reductions against \underline{SS} are more modest, from 75.41 to 74.68 and 74.26; while against \underline{MD}, the scores drop from 71.77 to 69.76 and 69.68. 
These differences stem from the unique matching strategies of each tracker. 
\underline{DS} prioritizes matching newly tracked objects, which can result in previously tracked objects being incorrectly matched with false alarms during an attack. Conversely, \underline{SS} and \underline{MD} treat all tracked objects equally, reducing the likelihood of incorrect ID assignments. 
\sys{}, capable of attacking any tracker that employs a feature bank, operates effectively regardless of a tracker’s specific matching procedures.
As shown in Table~\ref{tab:various_comb}, \sys{} is effective across various combinations of detectors and trackers with increased Mahalanobis thresholds. Additional experiments with three detectors (Faster R-CNN, FoveaBox, and DETR) with three trackers (\underline{DS}, \underline{SS}, and \underline{DT}) show \sys{}'s effectiveness across different architectures (two-stage, anchor-free, and transformer).

\begin{table}[t!]
\centering
\caption{Varying combinations of $R^{s}$ and $\epsilon$ with $R^{s}=10$.}
\label{tab:iteration}
\vspace{-0.3cm}
\resizebox{0.48\textwidth}{!}{%
\begin{tabular}{c|c|c|c|c|c|c|c||c|c}
\hline
  $\epsilon$ & \multicolumn{7}{c||}{4/255} & \multicolumn{2}{c}{16/255} \\ \hline
  $R^{s}$  & 20 & 40 & 60 & 80 & 100 & 120 & 140 & 10 & 20 \\ \hline
  IDF-1  & 68.5 & 65.34 & 63.65 & 64.63 & 63.2 & 63.84 & \textbf{63.32} & 67.6 & \textbf{59.4} \\ \hline
\end{tabular}
}
\end{table}

\begin{table}[t!]
\caption{Impact of Step 2 in \sys{}.}
\label{tab:wo_step2}
\vspace{-0.3cm}
\centering
\resizebox{0.48\textwidth}{!}{%
\begin{tabular}{c|ccc|ccc|ccc}
\hline
Method & \multicolumn{3}{c|}{Clean} & \multicolumn{3}{c|}{w/o Step 2} & \multicolumn{3}{c}{\textbf{\sys{}}} \\ \hline
Tracker              &\underline{DS} & \underline{SS} & \underline{MD} & \underline{DS} &\underline{SS} & \underline{MD} & \underline{DS} & \underline{SS} & \underline{MD} \\ \hline
IDF1          & 79.78 & 75.43 & 71.77 & 77.52 & 72.44 & 68.31& \textbf{58.01} & \textbf{61.84} & \textbf{57.75} \\ \hline
\end{tabular}
}
\vspace{-0.3cm}
\end{table}

\subsection{Ablation study}
\label{subsec:ablation}

\textbf{Varying Mahalanobis distance threshold.}
In feature-based matching, trackers check if detected objects are within the tracked objects' \textit{matching boundary}, set by a threshold $\lambda^{m}$. 
The Kalman filter models each tracked object's motion details (e.g., center, width, and height) using a chi-square distribution, represented by the probability density function $f(x)$. 
The matching boundary is where 95\% of position values are concentrated around the distribution's mean within a $\lambda^{m}$ distance. 
Figs.~\ref{fig:Mahalanobis_false_alarms}(a)--(c) show IDF1 variations as $\lambda^{m}$ increases (e.g., $\times 5$ indicates $\lambda^{m}\times 5$), noting that a higher $\lambda^{m}$ expands the matching boundary, with ``Max" occurring when $\int^{\lambda^{m}}_{0}f(x)dx = 1$. 
Clean and the FN attack experience minimal IDF1 changes with rising thresholds. 
Conversely, Daedalus and the F\&F attack encounter an IDF1 increase and a notable decline in attack success as the threshold grows. 
\sys{} shows a high enhancement of the attack's effectiveness at higher thresholds,
indicating its reliance on the matching boundary to facilitate ID switches.

\textbf{Quantity of allowable false alarms.}
To evaluate attack effectiveness against practicality, we limited the number of false positives each attacker could generate. Figs.~\ref{fig:Mahalanobis_false_alarms}(d)--(f) plot IDF1 against the maximum number of objects for \underline{DS}, \underline{SS}, and \underline{MD}, respectively, with $D^{GT}$ representing the actual object count per attack frame and $2D^{GT}$, $3D^{GT}$, and $4D^{GT}$ denoting multiples of this number. 
The F\&F attack typically generates four false positives per targeted object; instead of reducing these numbers, we constrained the targeted objects per frame. Clean and \sys{} do not generate additional objects, hence their IDF1 scores remain unaffected by the object limit. Daedalus and the F\&F attack, however, show a notable IDF1 decrease with more false positives, although less significant than with \sys{}.

\textbf{Impact of Step 2.} 
As detailed in Fig.~\ref{fig:combined}(c), Step 2 is crucial for \sys{}. 
Table~\ref{tab:wo_step2} shows that ID switches occur less without Step 2, resulting in a smaller IDF1 reduction. 
For \underline{DS}, omitting Step 2 results in a minor IDF1 drop (79.78 to 77.52), while including it reduces IDF1 by 21.77 points (79.78 to 58.01), underscoring Step 2's importance.

\begin{figure}[t!]
    \centering
    \includegraphics[width=1.0\linewidth]{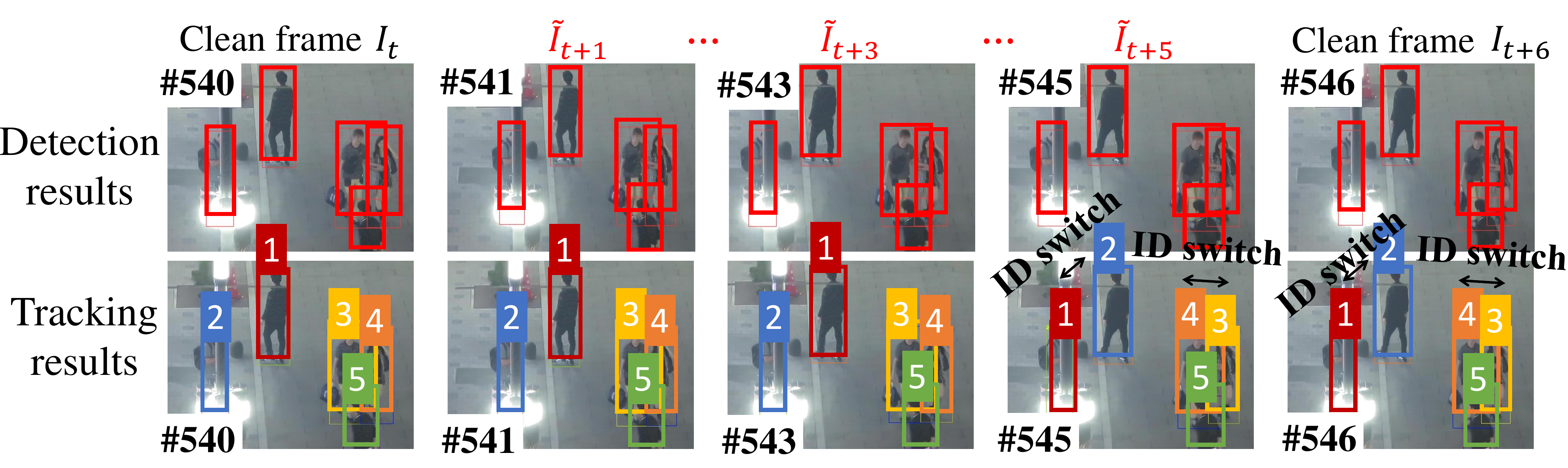}
    \caption{Visualized attack result of \sys{} on MOT17-04, which presents the detection results on the first line and the association outcomes on the second line, with tracking IDs represented by different colors.}
    \label{fig:visualization}
    \vspace{-0.3cm}
\end{figure}

\textbf{Number of iterations.}
Table.~\ref{tab:iteration} shows the effect of varying \sys{}'s $R^{s}$ on IDF1. 
Increasing $R^{s}$ leads to a greater reduction in IDF1, indicating that more iterations significantly enhance \sys{}'s attack efficiency.
Table~\ref{tab:iteration} shows that increasing $\epsilon$ (in Eqs.~\eqref{eq:pgd} and~\eqref{eq:iter_delta}) dramatically reduces $R^{s}$ needed to generate perturbations (which also reduces latency) while maintaining attack performance. 
For instance, with $\epsilon=4/255$ (default), accuracy drops from 77.4 to 63.3. With $\epsilon=16/255$, accuracy drops to 67.6 and 59.4 for $R^{s}=10$ and $R^{s}=20$, respectively.

\textbf{Visualization.}
Fig.~\ref{fig:visualization} visualizes how \sys{} can execute attacks while satisfying criteria efficiency and practicality on MOT17-04 (i.e., video number is 04 of MOT17). 
The detection results are on the first line, and the association outcomes are on the second line, with tracking IDs represented by different colors, where $``\#"$ represents the frame number.
As seen in Fig.~\ref{fig:visualization}, \sys{} changes objects' IDs during Step 2 ($\tilde{I}_{t+4}$ and $\tilde{I}_{t+5}$), and these ID changes persist in subsequent frames ($I_{t+6}$) even without further attacks, ensuring the IDs do not revert to their original states (efficiency holds). 
Additionally, since \sys{} does not perform adversarial attacks against the detector, the number of detected objects remains consistent before, during, and after the attack (practicality holds).

\section{Conclusion}

In this paper, we proposed a novel adversarial attack, \sys{}, designed to deceive multi-object trackers by attacking feature extractors in the association phase to trigger persistent ID switches. 
Our method is robust against heightened Mahalanobis distance thresholds and does not rely on false alarms or motion prediction for effectiveness. 
We demonstrated our approach's versatility by applying it to three popular multi-object trackers, DeepSORT, StrongSORT, and MOTDT with one-stage, two-stage, anchor-free, and transformer detectors. 
Through comprehensive experiments on public datasets, we explored a range of attacking tactics and established the effectiveness of \sys{}.

\begin{figure}[t!]
    \centering
    \includegraphics[width=1\linewidth]{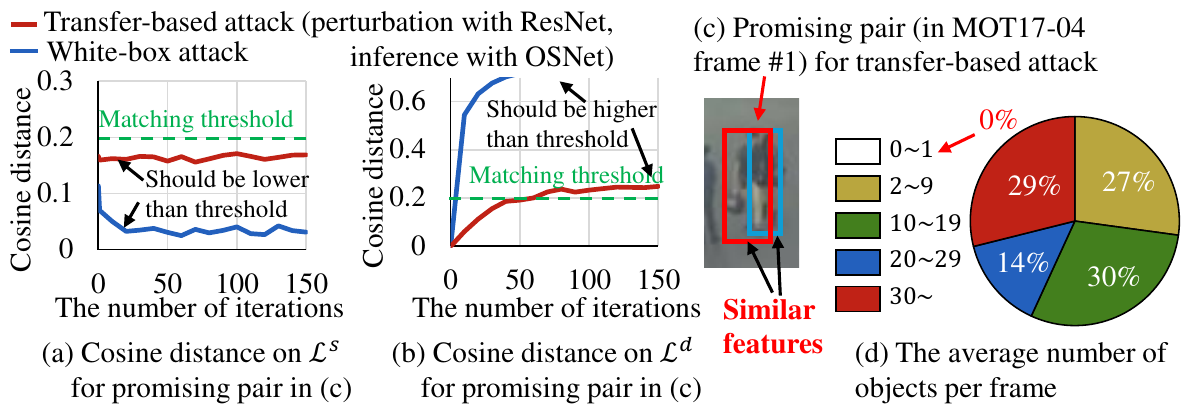}
    \vspace{-0.8cm}
    \caption{ (a)--(c) Experimental analysis of transfer-based attack for feature extractor and (d) the average number of objects in each frame.}
    \label{fig:feature_cost}
    \vspace{-0.3cm}
\end{figure}

\textbf{Limitation.} 
While there has been significant research on black box attacks in classification problems~\cite{MWZ23,LHZ22}, there has been no remarkable progress in attacks in the domain of multi-object detection (and tracking) due to its inherent challenge~\cite{DCY24}.
Consequently, an MOT attack based on black-box methodologies has yet to be proposed.
Although \sys{} is also based on a white-box attack, as illustrated in Fig.~\ref{fig:feature_cost}(c), a transfer-based black-box attack on \sys{} can be feasible when object pairs (A and B) for an ID switch have similar features. 
This is because generating similar features for A from B (or vice versa) is easier, and creating untargeted features (e.g., X and Y) is simpler than targeting generation. 
Figs.~\ref{fig:feature_cost}(a) and (b) show that, while less effective than a white-box attack, this approach meets the necessary thresholds for \sys{} in a transfer-based attack. 
Thus, a black-box attack will work if it finds object pairs with similar features to those of the deployed detector. 
Research on attacking MOT, targeting multiple objects simultaneously, is still in its early stages. 
Notable studies include the F\&F attack and Hijacking, both white-box attacks, while \sys{} significantly addresses their limitations. 
As \sys{} induces an ID switch between two target objects, it cannot switch its ID if there is only one object in the initial frame.  
However, Fig.~\ref{fig:feature_cost}(d) shows no frames with only one object; even the sparsest frames have at least two objects, which also holds for MOT20.

\appendix

\bigskip

\bibliography{aaai25}

\end{document}